\pdfoutput=1
\documentclass[11pt]{article}

\usepackage[]{acl}

\usepackage{times}
\usepackage{latexsym}

\usepackage[T1]{fontenc}
\usepackage[mathscr]{euscript}
\usepackage[utf8]{inputenc}

\usepackage{microtype}

\usepackage{booktabs,subcaption,amsmath}
\usepackage{tikz}
\usepackage{tikz-qtree}
\usepackage{enumitem} % compact item
\usepackage{fancyvrb}
\usepackage{mathtools}
\usepackage{amsmath}
\usepackage{multirow}
\usepackage{algorithm}
\usepackage[noend]{algpseudocode}
\usepackage{mathtools}
\usepackage{booktabs}
\usepackage{wrapfig}
\usepackage{inconsolata}

\title{Retrieve-and-Fill for Scenario-based Task-Oriented Semantic Parsing}

\author{Akshat Shrivastava \quad\quad Shrey Desai \quad\quad Anchit Gupta \quad\quad Ali Elkahky \\ \quad\quad \textbf{Aleksandr Livshits} \quad\quad \textbf{Alexander Zotov} \quad\quad \textbf{Ahmed Aly} \\
Meta \\
\tt{\{akshats, shreyd, anchit, alielk,} \\\texttt{alll, alexzotov, ahhegazy\}@fb.com}}

\begin{document}
\maketitle
\begin{abstract}
Task-oriented semantic parsing models have achieved strong results in recent years, but unfortunately do not strike an appealing balance between model size, runtime latency, and cross-domain generalizability. We tackle this problem by introducing scenario-based semantic parsing: a variant of the original task which first requires disambiguating an utterance's ``scenario'' (an intent-slot template with variable leaf spans) before generating its frame, complete with ontology and utterance tokens. This formulation enables us to isolate coarse-grained and fine-grained aspects of the task, each of which we solve with off-the-shelf neural modules, also optimizing for the axes outlined above. Concretely, we create a Retrieve-and-Fill (RAF) architecture comprised of (1) a retrieval module which ranks the best scenario given an utterance and (2) a filling module which imputes spans into the scenario to create the frame. Our model is modular, differentiable, interpretable, and allows us to garner extra supervision from scenarios. RAF achieves strong results in high-resource, low-resource, and multilingual settings, outperforming recent approaches by wide margins despite, using base pre-trained encoders, small sequence lengths, and parallel decoding.
\end{abstract}

\section{Introduction}

\begin{figure*}[t]
\centering
\includegraphics[scale=0.5]{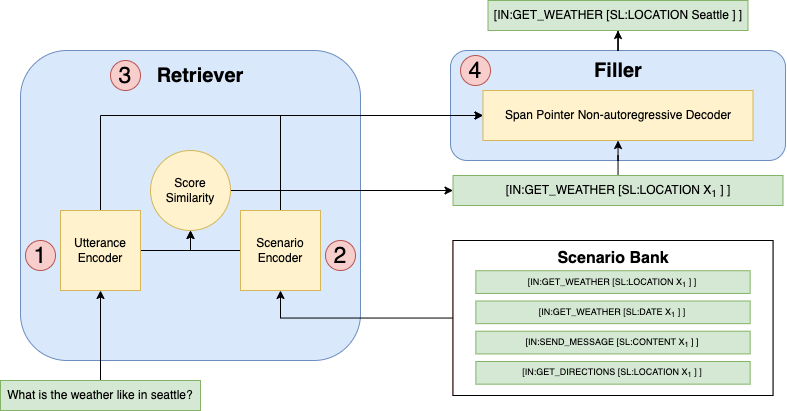}
\caption{\textbf{High-Level Overview.} Retrieve-and-Fill (RAF) consists of 4 steps: (1) Encode the utterance via an utterance encoder (e.g., RoBERTa); (2) Encode all scenarios \cite{desai2021lowresource} in the scenario bank into a cached index via a scenario encoder (e.g., RoBERTa); (3) Compute the dot product similarity amongst the incoming utterance and the index of all scenarios, obtaining the top-$n$ candidates; (4) For each retrieved scenario, leverage the non-autoregressive span pointer decoder \cite{spanpointernetwork} to impute each scenario's spans.}
\label{fig:span_pointer_model}
\end{figure*}

Task-oriented conversational assistants typically first use semantic parsers to map textual utterances into structured frames for language understanding \cite{hemphill1990atis,coucke2018snips,gupta2018semantic,rongali2020don,decoupled}. While these parsers achieve strong performance with rich supervision, they often face obstacles adapting to novel settings, especially ones with distinct semantics and scarce data. Recent approaches address this by improving parsers' data efficiency, such as by using pre-trained representations \cite{decoupled, rongali2020don}, optimizing loss functions \cite{chen-2020-topv2}, and supplying natural language prompts \cite{desai2021lowresource}. However, these approaches typically rely on larger models and longer contexts, impeding their applicability in real-world conversational assistants. Even though non-autoregressive models can alleviate some concerns \cite{nar_semantic_parsing,spanpointernetwork,zhu2020dont}, to the best of our knowledge, there exists no approach which strikes an appealing balance between model size, runtime latency, and cross-domain generalizability.

We begin tackling this problem by introducing \textbf{scenario-based task-oriented semantic parsing} which is more closely tied to how new task domains are developed. A slight variation on original semantic parsing, we are given access to all supported scenarios apriori and have to parse the utterance given this scenario bank. Here, a scenario is akin to a \textit{incomplete} frame; it is, precisely, an intent-slot template with variables as leaf spans (e.g., \texttt{IN:GET\_WEATHER [SL:LOCATION} $x_1$ \texttt{] ]}), indicating it maps to a family of linguistically similar utterances. As domain development and data-collection usually starts out with designing the set of supported scenarios, our intuition is that by explicitly giving the model access to this bank we can train it to more explicitly reason about scenarios and
improve performance especially in the low data regime.

Concretely, we propose RAF (Retrieve-and-Fill), a modular yet differentiable architecture for scenario-based task-oriented semantic parsing. Guided by the definition of our task, RAF proceeds in two steps: (1) given an utterance, a \textbf{retrieval} module finds the highest ranking scenario and (2) given the utterance and retrieved scenario, a \textbf{filling} module imputes spans into the scenario, creating the final frame. This approach requires no extra supervision despite performing auxiliary inference: utterances and frames are typically provided, and scenarios are obtained by stripping leaf text in frames. We design RAF to capitalize on the advantages of prior work but avoid their disadvantages; using base pre-trained encoders across-the-board, our retrieval module caches intrinsic representations during inference and our filling module non-autoregressively decodes leaf spans in a generalizable fashion.

We evaluate our approach in high-resource, low-resource, and multilingual settings using standard task-oriented semantic parsing datasets. RAF achieves 87.52\% EM on TOPv2 \cite{chen-2020-topv2} and 86.14\% EM on TOP \cite{gupta2018semantic}, outperforming recent autoregressive and non-autoregressive models. RAF also excels in weakly supervised settings: on TOPv2-DA \cite{desai2021lowresource}, we outperform Inventory \cite{desai2021lowresource} on 4 domains (alarm, music, timer, weather) despite using <128 token sequence lengths and 2-4x less parameters, on TOPv2-LR \cite{chen-2020-topv2}, we outperform BART \cite{lewis2019bart}, RINE \cite{RINE}, and RoBERTa + Span Pointer \cite{spanpointernetwork} by wide margins on weather, and on MTOP \cite{li-2020-mtop}, we outperform XLM-R + Span Pointer \cite{spanpointernetwork}, achieving 42\% EM averaged across en$\rightarrow$\{es, fr, de, hi, th\} transfer tasks.

To summarize, our contributions are: (1) Introducing scenario-based task-oriented semantic parsing, a novel task which requires disambiguating scenarios during typical utterance$\rightarrow$frame predictions; (2) Creating RAF (Retrieve-and-Fill), a modular yet differentiable architecture composed of a retrieval and filling module for solving scenario-based task-oriented semantic parsing, leveraging novel intrinsic representations of scenarios; and (3) Achieving strong results in high-resource, low-resource, and multilingual settings, outperforming recent models such as Span Pointer, Inventory, and RINE by large margins, while also optimizing for model size, runtime latency, and cross-domain generalizability.

\begin{table*}[]
\small
\setlength{\tabcolsep}{4pt}
\begin{tabular}{ll}
\toprule
\textbf{Utterance} & \textbf{Frame} \\
\midrule
\multicolumn{2}{l}{Scenario: \texttt{[IN:GET\_WEATHER [SL:LOCATION} $x_1$ \texttt{] ]}} \\
\midrule
what's the weather in seattle & \texttt{[IN:GET\_WEATHER [SL:LOCATION} seattle \texttt{] ]} \\
how's the forecast in sf & \texttt{[IN:GET\_WEATHER [SL:LOCATION} sf \texttt{] ]} \\
\midrule
\multicolumn{2}{l}{Scenario: \texttt{[IN:GET\_WEATHER [SL:LOCATION} $x_1$ \texttt{] [SL:DATE\_TIME} $x_2$ \texttt{] ]}} \\
\midrule
what's the weather in seattle tomorrow & \texttt{[IN:GET\_WEATHER [SL:LOCATION} seattle \texttt{] [SL:DATE\_TIME} tomorrow \texttt{] ]} \\
how's the forecast in sf at 8pm & \texttt{[IN:GET\_WEATHER [SL:LOCATION} sf \texttt{] [SL:DATE\_TIME} 8pm \texttt{] ]} \\
\bottomrule
\end{tabular}
\caption{\textbf{Scenario Description.} Scenarios are intent-slot templates with missing slot text, suggesting they subclass linguistically similar utterances. We show examples of utterances, scenarios, and frames in the weather domain; each scenario consists of multiple (utterance, frame) pairs.}
\label{tab:scenarios}
\end{table*}

\section{Scenario-based Semantic Parsing}
\label{sec:secnario_parsing}

We formally introduce the task of scenario-based semantic parsing. Task-oriented conversational assistants support a wide range of domains (e.g., calling, reminders, weather) to maximize coverage over users' needs. Over time, in response to requests and feedback, developers often iterate on assistants' skill-sets by adding new domains. Though the crowdsourcing process of collecting and annotating samples---utterances and frames---for new domains can be accomplished in many ways, we propose a two-step methodology where developers (a) develop scenarios which roughly describe a family of samples and (b) collect linguistically-varied samples consistent with each scenario. We elaborate more on our scenario-based methodology below.

\paragraph{Scenario Definition.} We define a \textbf{scenario} as an intent-slot rule which abstracts away linguistic variation in utterances. More specifically, it is a decoupled semantic frame \cite{decoupled} with variables in leaf spans, indicating it can subclass utterances with similar syntactic and semantic structure, an example is showing in table \ref{tab:scenarios}. 
Our notion of a scenario is inspired by production rules in constituency parsing \cite{chomsky}, but the parallel is not exact given our scenarios have semantic not syntactic types.

Using scenarios, we can effectively quantize the space of possible utterances by identifying and defining slices of user requests. Our solution also offers fine-grained control over precision and recall, which is important in real-world systems; we can collect more paraphrases to improve precision and we can create more scenarios to improve recall. %Taking these elements together, scenarios can enable hybrid rule- and neural-based modeling approaches, which we will discuss further in Section XXX.

\paragraph{Case Study: Weather.} Figure~\ref{tab:scenarios} shows an example setting where we crowdsource weather domain samples using the scenario-based methodology outlined above. To begin, we may envision building a weather system which supports requests with location and/or date-time information. Therefore, we can define two scenarios: (1) a family of samples with one location span; and (2) a family of samples with one location span and one date-time span. Each scenario explicitly outlines the intents and slots that must be present, as scenarios are not ``one-size-fits-all'' rules with optional slotting. So, as an example, the utterance ``how's the forecast in sf'' would not be compatible with the scenario \texttt{[IN:GET\_WEATHER [SL:LOCATION} $x_1$ \texttt{] [SL:DATE\_TIME} $x_2$ \texttt{] ]} since it does not have a date-time span as specified by the $x_2$ variable.

\paragraph{Generalization Types.} There are multiple types of generalization developers may care about when building conversational systems; we specifically focus on two, namely \textbf{intra-scenario} and \textbf{extra-scenario} generalization. By intra-scenario generalization, we refer to a systems' ability to parse linguistic variations of the same scenario, and by extra-scenario generalization, we refer to a systems' ability to parse scenarios with unseen intent-slot combinations. Using the weather example in Table~\ref{tab:scenarios}, a system trained on \textbf{the first scenario only} may exhibit the following behavior: with intra-scenario generalization, it would be able to parse utterances \textit{inside} the training scenario, and with extra-scenario, it would be able to parse utterances \textit{outside} the training scenario. \textbf{Our focus is primarily on intra-scenario generalization, as we argue high precision over an illustrative set of scenarios is more important in the early stages of domain development.} But, extra-scenario generalization is an interesting avenue of future work, as it offers opportunities to explore techniques with data augmentation \cite{geca}, inductive biases \cite{oren2020}, compositionality \cite{gupta2018semantic}, etc.

\paragraph{Task Definition.} Finally, we precisely define our task of scenario-based semantic parsing. For the typical task of semantic parsing, we define $U$ as a random variable over utterances, $F$ as a random variable over frames, and model $P(F|U)$: find the most likely frame given the utterance. However, because we introduce scenarios as a coarse, intermediate representation of frames, we additionally define $\mathscr{S}$ as the set of all supported scenarios given a priori, and model $P(F|U,\mathscr{S})$.

%\paragraph{Language Generalization} we define as a known semantic structure (intent, slot combination) however the underlying language is not known. For instance "whats the weather in seattle", "hows the weather in SF", "I am headed to LA, hows the forecast?". Each of these utterances is unique in its meaning, however all map to the same structure (intent get weather, slot location). Requiring our model to generalize across languages. Losely, we can view multilinguality as language scaling, as the goal is to support existing domains across various languages.
%\paragraph{Structural Generalization} we define as synthesizing new intent slot combinations for deeper features, an example here is compositional parsing. Where we want to be able to support combinations of various intents and slots. 
%With this in mind, we focus our problem to target language scaling, in early stages of domains, we believe it is easy for a domain developer to expand the necessary structures by themselves, and we want to minimize the amount of natural data needed to standup a feature.
%\subsection{Formal Problem Definition}
%Let $X$ be the input utterance, $Y$ is the semantic parse, traditional parsing models $P(Y|X)$ through various modeling techniques (seq2seq for instance). In scenarios based parsing, we are provided with a list of structure templates $T$, and we can leverage this template during decoding, removing the need for structural generalization and focusing on language generalization $P(Y | X, T)$.

\section{RAF: Retrieve and Fill}
\label{sec:RAF}

We propose a model called RAF (\underline{R}etrieve \underline{a}nd \underline{F}ill) for our scenario-based semantic parsing task that naturally decomposes the task into a coarse-to-fine objective where we (a) find the most likely scenario given the utterance and (b) find the most likely frame given the utterance and scenario. More concretely given an utterance $u$ and scenarios $s_1, \cdots, s_n$, as well as a gold scenario $s^*$ and gold frame $f^*$, we learn our model as follows:
\begin{enumerate}
    \item Retrieve (Coarse Step; \S \ref{sec:retrieval}): A \textbf{retrieval module} maximizes $P(S=s^*|U=u)$ by learning to retrieve scenario $s_i$ given utterance $u$, e.g., ``what's the weather in seattle'' $\rightarrow$  \texttt{[IN:GET\_WEATHER [SL:LOCATION} $x_1$ \texttt{] ]}.
    \item Fill (Fine Step; \S \ref{sec:filling}): A \textbf{filling module} maximizes $P(F=f^*|U=u,S=s^*)$ by decoding the most likely frame $f$ given the structure of scenario $s^*_i$ and spans of utterance $u$, e.g., \texttt{[IN:GET\_WEATHER [SL:LOCATION} $x_1$ \texttt{] ]} $\rightarrow$ \texttt{[IN:GET\_WEATHER [SL:LOCATION} seattle \texttt{] ]}.
\end{enumerate}

RAF is a composable yet differentiable model which implements coarse-to-fine processing: we first develop a coarse-grained sketch of an utterance's frame, then impute fine-grained details to achieve the final frame. In these types of approaches, there often exists a trade-off when creating the intermediate representation; if it is ``too coarse'', the filling module suffers, but if it is ``too fine'', the retrieval module suffers. We find scenarios, as defined in \S \ref{sec:secnario_parsing}, offer the most appealing solution, as the retrieval module unearths rough syntactic-semantic structure, while the filling module focuses in on imputing exact leaf spans. 

This objective is similar, in spirit, to other setups in syntactic and semantic parsing \cite{Dong2018CoarsetoFineDF}, but we explore different parameterizations of the likelihood functions. In the following sub-sections, we discuss the technical details behind the retrieval and filling modules, as well as describe the training and inference procedures.

%We propose a new modeling paradigm \textbf{RAF}: \textbf{R}etrieve \textbf{a}nd \textbf{F}ill, that focuses on decomposing semantic parsers for scenario based semantic parsing into 3 components: utterance encoder, frame encoder, and a decoder. Functionally this decomposition to frame retrieval, span decoding steps resembles a \textbf{coarse-2-fine} modeling scheme. INSERT MODEL DIAGRAM HERE. 
%We describe 3 criteria in building our model (1) Efficiency, the model should be built suited for realtime deployment (2) Scalability, minimize the amount of data needed to stand up this model (3) Interpretability, make sure we can understand our model predictions.
%In this section we describe in detail the components involved in our model.

\subsection{Retrieval Module}
\label{sec:retrieval}

First, we discuss the \textbf{coarse-grained step} of RAF, which aims to find the best-fitting scenario for an utterance. 

A typical solution is learning a classifier which explicitly maximizes the conditional probability $P(S=s^* | U=u)$ of an utterance $u$ mapping to a scenario $s^*$. The classifier can be parameterized with a pre-trained encoder, and its parameters can be learned via maximum likelihood estimation. However, the multi-class classification approach has a couple of disadvantages; each scenario $s^*$ contains useful signal in its structure (e.g., its intents and slots) which is not utilized and the output space itself cannot be ``hot swapped'' during inference, such as if new scenarios were to be dynamically added.

An alternative is formulating this task as a metric learning problem. Here, we can maximize the similarity $\mathrm{sim}(u, s^*)$ between an utterance $u$ and scenario $s^*$ as judged by a scalar metric. This offers numerous advantages: we can explore ad-hoc encodings of utterances and scenarios, adjust the output space dynamically during inference, and compute exact conditional probabilities by leveraging a (tractable) partition function.

\subsubsection{Bi-Encoder Retrieval}
\label{sec:retrieval-biencoder}

Following retrieval modeling in open-domain QA \cite{dpr}, we specifically leverage pre-trained encoders ($E_U$ for utterances and $E_S$ for scenarios) to compute dense vector representations, then maximize the dot product similarity $\mathrm{sim}(u, s^*) = E_U(u)^{\top}E_S(R(s^*))$ between utterance-scenario pairs $(u, s^*)$; the precise nature of $R$ is discussed in \S \ref{sec:frame_representation}. To learn such a metric space and avoid degenerate solutions we need to train the encoders to \textit{pull} positive pairs together and \textit{push} negative pairs apart. Hence, we need access to both positive (gold; $(u, s^+)$) and negative (non-gold; $(u, s^-)$) pairs. 

% As such following \cite{simclr} we define the constrastive loss for a single example tuple as $(u_i, s^+_i, s^-_i)$:

% \begin{multline}
% \small
% \mathcal{L}_\textrm{retrieval}(u_i, s^+_i, s^-_i) = \\
% \log \frac{\exp(\mathrm{sim}(u_i, s^+_i))}{\exp(\mathrm{sim}(u_i, s^+_i)) + \exp(\mathrm{sim}(u_i, s^-_i))}
% \label{eq:single-neg-loss}
% \end{multline}

\subsubsection{Negatives Sampling}
\label{sec:negatives}

The choice of negatives has a large impact on retrieval performance, which is consistent with findings in information retrieval \cite{zhan2021optimizing, dpr}. We explore two types of negative sampling to improve retrieval performance: in-batch negatives and model-based negatives.

\paragraph{In-Batch Negatives.} We mine positive and negative pairs from each training batch using in-batch negatives \cite{dpr}. Let $\mathbf{U}$ and $\mathbf{S}$ be the utterance and scenario matrices, each being a $(B \times d)$ matrix consisting of $d$-dimensional embeddings up to batch size $B$. We obtain $B^2$ similarity scores upon computing the similarity matrix $\mathbf{M} = \mathbf{U}\mathbf{S}^\top$, where $\mathbf{M}_{i=j}$ consists of positive scores and $\mathbf{M}_{i\ne j}$ consists of negative scores. For each positive utterance-scenario pair $(u_i, s^+_i)$, we now have $(B-1)$ negative pairs $\{(u_i, s^-_{ij})\}_{i \ne j}$. Having collected multiple negative pairs per positive pair, we leverage the contrastive loss defined in \citet{dpr, simclr}: % for the single example tuple $(u_i, s^+_i, s^-_{i1}, \cdots, s^-_{im})$:

\begin{multline}
\mathcal{L}_\textrm{retrieval}(u_i, s^+_i, s^-_{i1}, \cdots, s^-_{im}) = \\
\log \frac{\exp(\mathrm{sim}(u_i, s^+_i))}{\exp(\mathrm{sim}(u_i, s^+_i)) + \sum^m_{i \ne j} \exp(\mathrm{sim}(u_i, s^-_{ij}))}
\end{multline}

\paragraph{Model-Based Negatives.} While in-batch negatives greatly expands the number of negatives, these negatives may not be particularly challenging as they are randomly sampled from the training dataset. To increase the quality of our utterance-scenario metric, we explore augmenting in-batch negatives with model-based negatives. Specifically, given a retrieval module from training round $t$ with metric $\mathrm{sim}^t(u, s) = E^t_U(u)^{\top}E^t_S(s)$, we find the top-$k$ neighbors for each positive pair $(u_i, s^+_i)$ by computing $\mathrm{argmax}_{(1\ldots k)} \{\mathrm{sim}^t(u_i, s^-_k) | 1 \le k \le n \land i \ne k \}$. Using this algorithm, we cache each training example's hard negatives, then in training round $(t + 1)$, we fine-tune the retrieval module using both in-batch and model-based negatives. Each non-seed training round $t > 1$ therefore features \textbf{at most} $(B(k + 1) - 1)$ negatives per positive pair. This procedure is similar to iterative training used in \cite{dprpaq}.

\paragraph{Identity Masking.} The precise number of negatives is empirically slightly less, as sometimes we see conflicts between each training examples' negative pairs. Let $(u_i, s_i)$ and $(u_j, s_j)$ be two training examples within the same batch. During model-based negatives sampling in training round $t$, $(u_j, s_j)$ may include $s_i$ as a top-$k$ negative pair. %: $s^+_i \in \{s^-_{j1}, \cdots, s^-_{jm}\}$. Then, in training round $(t + 1)$, due to the use of in-batch negatives, $(u_j, s_j)$'s negatives get inculcated into $(u_i, s_i)$'s negatives: $\{s^-_{j1}, \cdots, s^-_{jm}\} \subseteq \{s^-_{i1}, \cdots, s^-_{im}\}$. 
This complicates metric learning as $s_i$ becomes a positive \textit{and} negative for $u_i$ simultaneously. We therefore implement an identity mask on each training example's negatives which ensures no conflicts when mixing in-batch and model-based negatives.

\subsubsection{Scenario Representation}
\label{sec:frame_representation}

While we have covered scenario-based retrieval above, we have not yet precisely described how dense vectors for scenarios are computed. Recall our definition of utterance-scenario similarity in \S \ref{sec:retrieval-biencoder}: our objective is to maximize $\mathrm{sim}(u, s) = E_U(u)^\top E_S(R(s))$, where $R$ is a string transformation applied to scenarios.

Because we parameterize $E_U$ and $E_S$ with standard pre-trained encoders (e.g., RoBERTa), it follows that $E_U(u)$ and $E_S(s)$ must return approximately similar vectors in order to have a high dot product. However, a task setup using vanilla utterances and scenarios does not achieve this. Consider the utterance ``what's the weather in seattle'' and scenario \texttt{IN:GET\_WEATHER [SL:LOCATION} $x_1$ \texttt{] ]}. Pre-trained encoders can return strong return representations for the utterance, but not necessarily for the scenario; because they have been pre-trained on naturally-occurring, large-scale text corpora, they are unlikely to glean as much value from dataset-specific concepts, such as \texttt{IN:GET\_WEATHER} and \texttt{SL:LOCATION}.

An effective solution to this problem, as proposed by 
\citet{desai2021lowresource}, leverages \textbf{intrinsic modeling} to rewrite intents and slots as a composition of their intrinsic parts in a single string. \citet{desai2021lowresource} define two, in particular: the categorical type (e.g., ``intent'' or ``slot'') and language span (e.g., ``get weather'' or ``location''). Using this methodology, we can transform the scenario \texttt{IN:GET\_WEATHER [SL:LOCATION} $x_1$ \texttt{] ]} $\rightarrow$ \texttt{[ intent | get weather [ slot  | location} $x_1$ \texttt{] ]}, which is inherently more natural and descriptive.

Guided by the general concept of intrinsic modeling, our goal here is to define $R$ such that $E_U(u)^\top E_S(s) \le E_U(u)^\top E_S(R(s))$, all else being equal. We discuss our approach in detail in the following sub-sections.

\subsubsection{Language Spans}

\citet{desai2021lowresource} chiefly use an \textbf{automatic} method to extract language spans from ontology labels. For example, using standard string processing functions, we can extract ``get weather'' from \texttt{IN:GET\_WEATHER}. While this method is a surprisingly strong baseline, it heavily relies on a third-party developers' notion of ontology nomenclature, which may not always be pragmatically useful. In TOPv2 \cite{chen-2020-topv2}, our principal evaluation dataset, there exists ambiguous labels like \texttt{SL:AGE}---is this referring to the age of a person, place, or thing?

Therefore, to improve consistency and descriptiveness, we propose a \textbf{handmade} method where we manually design language spans for each ontology label.\footnote{See Appendix \ref{appendix:intrinsic_descriptions} for our curated intent and slot descriptions, respectively.} This approach is similar to prompting \cite{gpt3,prompting1,prompting2,prompting3}, but unlike typical prompt-engineering methods, we attempt to assign intuitive descriptions with no trial-and-error. While it is certainly possible to achieve better results with state-of-the-art methods, our aim is to be as generalizable as possible while still reaping the benefits of descriptive prompting.

The most frequent techniques we use are (1) using the label as-is (e.g., \texttt{IN:GET\_WEATHER} $\rightarrow$ ``get weather''); (2) inserting or rearranging prepositions (e.g., \texttt{IN:ADD\_TO\_PLAYLIST\_MUSIC} $\rightarrow$ ``add music to playlist''; and (3) elaborating using domain knowledge (e.g., \texttt{IN:UNSUPPORTED\_ALARM} $\rightarrow$ ``unsupported alarm request'').

\subsubsection{Example Priming}

Despite using curation to improve ontology label descriptions, there are still many labels which remain ambiguous. One such example is \texttt{SL:SOURCE}; this could refer to a travel source or messaging source, but without seeing its exact manifestations, it is challenging to fully grasp its meaning. This motivates us to explore \textbf{example priming}: augmenting scenario representations with randomly sampled, dataset-specific slot values. This can help our model further narrow down the set of spans each slot maps to during parsing. Furthermore, our examples are just spans, so they are straightforward to incorporate into our representation. For example, for the slot \texttt{SL:WEATHER\_TEMPERATURE\_UNIT}, we can augment and contextualize its representation ``slot | unit of weather temperature'' with ``slot | unit of weather temperature | F / C'' where ``F'' and ``C'' are examples which appear in our dataset.

\subsubsection{Representation Sampling}
\label{sec:representation_sampling}

Our scenario-based retrieval task performs reasoning over a compact set of scenarios, unlike large-scale, open-domain tasks such as information retrieval and question answering which inculcate millions of documents. As such, \textbf{the chance our system overfits to a particular representation is much greater}, no matter what it is set to. So, we instead make $R$ stochastic by uniformly sampling unique frame representations for encoding scenarios. We re-define $R$ as a discrete random variable with outcomes $\{r_1, \cdots, r_n\}$ where $P(R=r_i) = \frac{1}{n}$. Each $r_i$ denotes a unique scenario representation (e.g., a string using \textit{curated} descriptions \textit{without} examples); Table~\ref{tab:repr-types} enumerates the complete set of outcomes. While we intend to improve the generalizability of our system during training, we typically want deterministic behavior during testing, and therefore we restrict the outcomes to a singleton $\{r_i\}$.

\subsection{Filling Module}
\label{sec:filling}

Next, we discuss the \textbf{fine-grained} step of RAF, which aims to infill a retrieved scenario with utterance spans in leaf slots. Unlike the coarse-grained step, the fine-grained step can be modeled with off-the-shelf models from prior work in task-oriented semantic parsing.

This step is typically modeled as a seq2seq problem where a decoder maps a scenario $s^*$ to a frame $f^*$ using cross-attention on an utterance $u$, which results in the objective $P(F=f^* | U=u, S=s^*)$. Because frame $f^*$ is a sequence, this objective can be factorized in multiple ways: an \textbf{autoregressive} approach $\prod^n_{i=1} P(f^*_i|f^*_{<i-1},u,s^*)$ assumes conditional dependence on prior frame tokens while a \textbf{non-autoregressive} approach $\prod^n_{i=1} P(f^*_i|u, s^*)$ assumes conditional independence. We elect to use the latter given its recent successes in task-oriented semantic parsing; we refer interested readers to \citet{nar_semantic_parsing, spanpointernetwork} for technical details. We largely use their model as-is, but one modification we make is \textbf{scenario fusion}. Rather than the decoder learning an embedding matrix for scenario tokens from scratch, we initialize the decoder's embeddings with the scenario encoder's final state. Our final objective is $\mathcal{L}_\textrm{filling} = \mathrm{NLL}(f^*, f) + \alpha \mathrm{LS}(f)$ where $\mathrm{LS}$ refers to label smoothing \cite{pereyra2017regularizing}.

%\begin{itemize}
%    \item Prior sections discussed the first step of RAF "retrieve" here we focus on the second step "fill". The task for filling is defined as identifying slot values to fill in for the semantic frames (e.g. [IN:CREATE\_CALL [SL:\_CONTACT ] ] for call john we want to fill in John for the contact). 
%    \item To fill in the spans efficiently we rely on the span based decoding formulation proposed in \citet{SpanPointerNetwork}. This allows us to model filling as a span selection task through a non-autoregressive decoder leading to our decoding being efficient as well
%    \item We initialize decoder embeddings with embeddings retrieved from the frame encoder for each token and ask the model to specify the span indexes for the slot text, the model is optimized with the standard CE loss over the span pointer form.
%    \item The loss of this model is learned in a multi-task learning loss along side the retrieval loss described in the prior seciton.
%\end{itemize}

\subsection{Training and Inference}

Once we combine the coarse-grained and fine-grained steps, we now have an end-to-end system which maps an utterance (``get weather in seattle'') to a frame with ontology tokens (intents and slots) and utterance spans (e.g., \texttt{[IN:GET\_WEATHER [SL:LOCATION} seattle \texttt{] ]}).

Because our system requires strong negatives for accurate retrieval, we decompose training into three steps: (1) Train RAF using in-batch negatives only; (2) Sample the top-$n$ frames as model-based negatives from RAF in Step (1) by running it over the training set; (3) Train RAF using both in-batch negatives and model-based negatives. RAF requires three sets of parameters to train: the retrieval module consists of an utterance encoder $\theta_U$ and scenario encoder $\theta_S$, while the filling module consists of a frame decoder $\theta_F$. Our system is modularized and differentiable, therefore we fine-tune these modules using a joint loss $\mathcal{L}(\theta_U, \theta_S, \theta_F) = \beta \mathcal{L}_\textrm{retrieval}(\theta_U, \theta_S) + \mathcal{L}_\textrm{filling}(\theta_U, \theta_S,\theta_F)$ where $\beta$ is a scalar weighting term. In the low-resource setting We further augment our loss with an optional R3F term \cite{RXF} to encourage robust representations following \citet{spanpointernetwork}. Hyper parameter details are described in Appendix \S \ref{appendix:hyperparameters}.

During inference, we can pipeline RAF's components: given an \textbf{utterance} $u$, the retrieval module finds the best \textbf{scenario} $\mathbf{s}' = \mathrm{argmax}_{\mathbf{s}'} P(\mathbf{s}|u)$, and using the predicted scenario as a template, the filling module outputs a \textbf{frame} $\mathbf{f}' = \mathrm{argmax}_{\mathbf{f}'} P(\mathbf{f}|u,\mathbf{s}')$ with ontology tokens (intents and slots) and utterance spans. Importantly, to maintain inference efficiency, the scenario encoders' embeddings over the scenario set can be cached, since these are known a priori.

\section{Experiments and Results}

We evaluate RAF in three settings: a high-resource setting (100,000+ training samples), low-resource setting (1-1,000 training samples), and multilingual setting (0 training samples). Our goal here is to show that our system both achieves competitive performance on established benchmarks and offers substantial benefits in resource-constrained environments where training samples are limited.

\subsection{Datasets for Evaluation}

Following prior work in task-oriented semantic parsing \citet{nar_semantic_parsing, decoupled, spanpointernetwork, RINE, desai-2021-dianosing, desai2021lowresource, RetroNLU, rongali2020don}, we use 5 datasets for evaluation: TOP \cite{gupta2018semantic}, TOPv2 \cite{chen-2020-topv2}, TOPv2-LR (Low Resource; \citet{chen-2020-topv2}), TOPv2-DA (Domain Adaptation; \citet{desai2021lowresource}), and MTOP \cite{li-2020-mtop}. TOP and TOPv2 are used for high-resource experiments, TOPv2-LR and TOPv2-DA are used for low-resource experiments, and MTOP is used in multilingual experiments. 

%TOP and TOP evaluates parsers' abilities to produce nested frames across the vent and navigation domains. In comparison, TOPv2 consists of both linear and nested frames and extends TOP to the alarm, messaging, music, navigation, timer, and weather domains. TOPv2-LR and TOPv2-DA create few-shot splits using various sampling algorithms (cite). Finally, while TOP and TOPv2 consist entirely of utterances in English, MTOP provides gold translations in Spanish, French, German, Hindi, and Thai, making it useful for evaluating parsers in multilingual settings.

\subsection{Systems for Comparison}

We compare against multiple task-oriented semantic parsing models, which cover autoregressive (AR), and non-autoregressive (NAR) training. See \citet{decoupled, RINE, nar_semantic_parsing, spanpointernetwork} for detailed descriptions of these models.

The autoregressive models consist of BART \cite{lewis2019bart} and RoBERTa \cite{roberta}, and RINE \cite{RINE}, and the non-autoregressive models are RoBERTa NAR \cite{nar_semantic_parsing} and RoBERTa NAR + Span Pointer \cite{spanpointernetwork}. These models are applicable to both high-resource and low-resource settings; though, for the latter, we also add baselines from \citet{desai2021lowresource}: CopyGen (BART + copy-gen decoder) and Inventory (BART + intrinsic modeling). The multilingual setting only requires swapping RoBERTa with XLM-R \cite{conneau-2020-xlmr}.

We denote our system as RAF in our experiments. Unless noted otherwise, we use RoBERTa$_\textrm{BASE}$ for the utterance encoder $\theta_U$ and secnario $\theta_S$ and a random-init, copy-gen, transformer decoder for the frame decoder $\theta_F$. As alluded to before, we swap RoBERTa$_\textrm{BASE}$ with XLM-R$_\textrm{BASE}$ for multilingual experiments.

\begin{table}[t]
\centering
\setlength{\tabcolsep}{4pt}
\begin{tabular}{lrr}
    \toprule
    Model & TOPv2 & TOP \\
    \midrule
    \multicolumn{3}{l}{Type: Autoregressive Modeling (Prior)} \\
    \midrule
    RoBERTa$_\textrm{BASE}$ & 86.62 & 83.17 \\
    RoBERTa$_\textrm{LARGE}$ &  86.25  & 82.24  \\
    BART$_\textrm{BASE}$ & 86.73 & 84.33  \\
    BART$_\textrm{LARGE}$ & 87.48 & 85.71 \\
    \midrule
    \multicolumn{3}{l}{Type: Non-Autoregressive Modeling (Prior)} \\
    \midrule
    RoBERTa$_\textrm{BASE}$ & 85.78 & 82.37  \\
    \quad+ Span Pointer & 86.93 & 84.45  \\
    RoBERTa$_\textrm{LARGE}$ & 86.25 & 83.40 \\
    \quad+ Span Pointer & 87.37 & 85.07  \\
    \midrule
    \multicolumn{3}{l}{Type: Scenario Modeling (Ours)} \\
    \midrule
    RAF$_\textrm{BASE}$ & 87.14  & 86.00 \\
    RAF$_\textrm{LARGE}$ & \textbf{87.52}  & \textbf{86.14} \\
    \bottomrule
\end{tabular}
\caption{\textbf{High-Resource Results.} Exact Match (EM) on TOPv2 \cite{chen-2020-topv2} and TOP \cite{gupta2018semantic}. We compare various semantic parsing paradigms: autoregressive, non-autoregressive, and scenario. RAF achieves strong performance on TOPv2 and TOP, illustrating its competitiveness with state-of-the-art models.}
\label{tab:high-resource-results}
\end{table}

\begin{table}[]
\centering
\setlength{\tabcolsep}{3pt}
\begin{tabular}{lrrrr}
\toprule
 & alarm & music & timer & weather \\
\midrule
CopyGen$_\textrm{BASE}$ & 47.24 & 25.58 & 16.62 & 47.24 \\
CopyGen$_\textrm{LARGE}$ & 36.91 & 23.84 & 32.64 & 53.08 \\
Inventory$_\textrm{BASE}$ & 62.13 & 23.00 & 28.92 & 54.53 \\
Inventory$_\textrm{LARGE}$ & \textbf{67.25} & \textbf{38.68} & 48.45 & \textbf{61.77} \\
RAF$_\textrm{BASE}$ (ours) & 62.71 & 35.47 & \textbf{55.06} & 61.05 \\
\bottomrule
\end{tabular}
\caption{\textbf{High-Difficulty Low-Resource Results.} EM on the 1 SPIS split of TOPv2-DA \cite{desai2021lowresource}. Compared to Inventory and CopyGen baselines, RAF achieves competitive performance with a fraction of parameter usage.}
\label{tab:low-resource-results1}
\end{table}

\begin{table*}[t]
\centering
\setlength{\tabcolsep}{4pt}
\begin{tabular}{lrrrrrr}
\toprule
 & \multicolumn{6}{c}{Weather Domain (SPIS)} \\
\cmidrule(lr){2-7}
 & 10 & 25 & 50 & 100 & 500 & 1000 \\
\midrule
\multicolumn{7}{l}{Type: Autoregressive Modeling (Prior)} \\
\midrule
RoBERTa$_\textrm{BASE}$ AR & 69.71 & 74.90 & 77.02 & 78.69 & --- & 86.36 \\
BART$_\textrm{BASE}$ AR & 73.34 & 73.35 & 76.58 & 79.16 & --- & 86.25 \\
RINE$_\textrm{BASE}$ & --- & 74.53 & --- & --- & 87.80 & ---\\
RINE$_\textrm{LARGE}$ & --- & 77.03 & --- & --- & 87.50 & --- \\
\midrule
\multicolumn{7}{l}{Type: Non-Autoregressive Modeling (Prior)}  \\
\midrule
RoBERTa$_\textrm{BASE}$ NAR & 59.01 & 72.12 & 73.41 & 78.48 & --- & 87.42 \\
\quad+ Span Pointer & 72.03 & 74.74 & 74.85 & 78.14 & --- & \textbf{88.47} \\
\midrule
\multicolumn{7}{l}{Type: Scenario Modeling (Ours)} \\
\midrule
RAF$_\textrm{BASE}$ & \textbf{75.10} & \textbf{78.74} & \textbf{77.53} & \textbf{79.67} & \textbf{87.91} & 88.17 \\
\bottomrule
\end{tabular}
\caption{\textbf{Medium-Difficulty Low-Resource Results.} EM on various SPIS splits of the TOPv2 \cite{chen-2020-topv2} weather domain. RAF largely outperforms autoregressive and non-autoregressive models, trailing RoBERTa-Base + Span Pointer only in a high-resource split.}
\label{tab:low-resource-results2}
\end{table*}

\begin{table*}[h]
\centering
\begin{tabular}{lrrrrrrr}
\toprule
 & \multicolumn{7}{c}{Zero-Shot Evaluation} \\
\cmidrule(lr){3-8}
 & \texttt{en} & \texttt{en$\rightarrow$es} & \texttt{en$\rightarrow$fr} & \texttt{en$\rightarrow$de} & \texttt{en$\rightarrow$hi} & \texttt{en$\rightarrow$th} & Avg \\
\midrule
XLM-R$_\textrm{BASE}$ NAR & 78.3 & 35.2 & 32.2 & 23.6 & 18.1 & 16.7 & 25.2 \\
\quad + Span Pointer & \textbf{83.0} & 51.2 & 51.4 & \textbf{42.0} & 29.6 & 27.3 & 40.3 \\
% XLM-R$_\textrm{LARGE}$ NAR & 80.5 & 50.9 & 51.5 & 38.7 & 31.6 & 22.8 & 39.1 \\
% \quad + Span Pointer & \textbf{84.5} & \textbf{60.4} & \textbf{63.1} & \textbf{56.2} & \textbf{41.2} & \textbf{41.7} & \textbf{52.5} \\
% XLM-R$_\textrm{LARGE}$ AR$^\diamondsuit$ & 83.9 & 50.3 & 43.9 & 42.3 & 30.9 & 26.7 & 38.8 \\
RAF$_\textrm{BASE}$ (ours) & 81.1 & \textbf{56.0} & \textbf{54.9} & 40.1 & \textbf{32.1} & \textbf{30.0} & \textbf{42.6} \\
\bottomrule
\end{tabular}
\caption{\textbf{Multilingual Results.} We perform zero-shot experiments where we fine-tune a parser on English (\texttt{en}), then evaluate it a non-English language---Spanish (\texttt{es}), French (\texttt{fr}), German (\texttt{de}), Hindi (\texttt{hi}), and Thai (\texttt{th})---without fine-tuning. Average EM (Avg) is taken over the five non-English languages. RAF outperforms XLM-R-Base + Span Pointer by +2.3\% on average.}
\label{tab:multilingual-results}
\end{table*}

\subsection{High-Resource Setting}

First, we evaluate RAF in a high-resource setting where hundreds of thousands are samples are available for supervised training; Table~\ref{tab:high-resource-results} shows the results. RAF achieves strong results across-the-board, using both base and large pre-trained encoders: \textbf{RAF$_\textrm{BASE}$ consistently outperforms other base variants by 0.25-0.5 EM and RAF$_\textrm{LARGE}$ comparatively achieves the best results on TOP and TOPv2.}

\subsection{Low-Resource Setting}

Having established our system is competitive in high-resource settings, we now turn towards evaluating it in low-resource settings, where training samples are not as readily available. Here, we chiefly consider two setting types: a \textbf{high difficulty} setting (TOPv2-DA) with 1-10 samples and a \textbf{medium difficulty} setting (TOPv2-LR) with 100-1,000 samples. The exact number of samples in a few-shot training subset depend on both the subset's cardinality and sampling algorithm.

Tables \ref{tab:low-resource-results1} and \ref{tab:low-resource-results2} show results on the high and medium difficulty settings, respectively. \textbf{RAF achieves competitive results in the high-difficulty setting, outperforming both CopyGen$_\textrm{BASE}$ and Inventory$_\textrm{BASE}$ by large margins}; notably, on timer, we nearly double Inventory$_\textrm{BASE}$'s exact match score. \textbf{RAF also performs well in the medium-difficulty setting; our system consistently outperforms prior autoregressive, and non-autoregressive models.} These results particularly highlight the effectiveness of coarse-to-fine modeling: our retrieval module learns a generalizable notion of utterance$\rightarrow$scenario mappings, especially given that utterance and scenarios are scored on semantic alignment, and our filling module precisely infills utterance spans into the scenario template.

\subsection{Multilingual Setting}

Finally, we consider a multilingual setting, where a model trained on English samples undergoes zero-shot transfer to non-English samples. In Table~\ref{tab:multilingual-results}, we see that, \textbf{compared to XLM-R$_\textrm{BASE}$ NAR + Span Pointer, RAF achieves +2.3 EM averaged across all 5 non-English languages.} Upon inspecting this result more closely, RAF's performance is strong across both typologically \textit{similar} languages (+4.8 EM on Spanish, +3.5 EM on French) and \textit{distinct} languages (+2.7 EM on Hindi and Thai). A key reason for RAF's strong performance is most of the domain shift is localized to the retrieval module. In MTOP, utterances across multiple languages have linguistic variation but their scenarios are the exact same, and so a bulk of our system's work is in retrieving the correct scenario. While our results illustrate the efficacy of a monolingual retriever, we can explore creating a multilingual retriever in future work; due to our system's modularity, such a retriever can simply be a drop-in replacement for the current one.

\section{Ablations and Analysis}

While our experiments show RAF achieves strong performance in high-resource, low-resource, and multilingual settings, we have not yet isolated the contribution of its different components. Here, we perform several experiments to better understand RAF's design decisions.

\subsection{Model Ablations}
\label{sec:model_ablation}

\begin{table}[t]
\centering
\setlength{\tabcolsep}{4pt}
\begin{tabular}{lrr}
\toprule
Model & EM & EM-S \\
\midrule
Classify and Fill & 84.80 & 87.16 \\
RAF$_\textrm{BASE}$ & \textbf{87.03} & \textbf{89.34} \\
\quad - Hard Negatives & 83.69 & 86.00 \\
\quad - Identity Masking & 85.87 & 88.23 \\
\quad - Scenario Fusion & 86.26 & 88.69 \\
\quad - Parameter sharing & 86.76 & 89.10 \\
\quad - Repr. Sampling & 86.70 & 89.05 \\
\quad + Heuristic negatives  & 85.66 & 89.10 \\ 
%\quad + Gold scenario & \textbf{96.56} & \textbf{100.00} \\
\bottomrule
\end{tabular}
\caption{\textbf{Model Ablations.} We assess several components of our model, individually removing them and evaluating EM (Exact Match) and EM-S (Exact Match of Scenarios, i.e., intent-slot templates without slot text) on the TOPv2 \cite{chen-2020-topv2} validation set.}
\label{tab:model_ablation}
\end{table}

\begin{table*}[t]
\centering
\setlength{\tabcolsep}{4pt}
\begin{tabular}{lrrrr}
\toprule
& \multicolumn{2}{c}{Source Fine-Tuning} & \multicolumn{2}{c}{Target Fine-Tuning} \\
\cmidrule(lr){2-3} \cmidrule(lr){4-5}
Model & EM & EM-S & EM & EM-S \\
\midrule
Canonical Repr. & 86.67 & 88.74 & 74.25 & 76.80 \\
Intrinsic Repr. & \textbf{87.10} & \textbf{89.15} & 78.74 & 81.70 \\
\quad - Automatic & 87.02 & 89.06 & 78.92 & 81.80 \\
\quad - Handmade & 86.85 & 88.87 & \textbf{79.27} & \textbf{82.29} \\
\quad - Examples & 86.91 & 88.90 & 78.25 & 80.89 \\
\bottomrule
\end{tabular}
\caption{Scenario representation ablation, where the target domain is a 25 SPIS split of the TOPv2 \cite{chen-2020-topv2} weather domain. Following the typical few-shot fine-tuning methodology, we perform source fine-tuning on all domains except weather and reminder, then perform target fine-tuning on a specific split. Encouragingly, we see intrinsic representations result in better EM and EM-S.}
\label{tab:intrinsic_ablation}
\end{table*}

First, we perform model ablations on RAF, removing core retrieval- and filling-related components we originally introduced in \S \ref{sec:RAF}. From Table \ref{tab:model_ablation}, we draw the following conclusions:

\paragraph{Negatives are important for accurate scenario retrieval.} The metric learning objective for retrieval, as introduced in \S \ref{sec:negatives}, precisely delineates between positive and negative samples. Our ablations show model-based negatives and identity masking are critical to achieve best performance; when removing model-based negatives, for example, retrieval accuracy drops by 3\%+. We also investigate training RAF with heuristic-based negatives: a simple algorithm which finds top-$k$ similar scenarios with string-based edit distance (\S \ref{appendix:heuristic_negative}). However, heuristic-based negatives regress both retrieval-only and end-to-end approach, suggesting model-based negatives are more informative.

\paragraph{Sharing parameters between retrieval encoders improves quality.} Our retrieval module has two encoders: an utterance encoder $E_U$ and a scenario encoder $E_S$. An important design decision we make is tying both encoders' parameters together with RoBERTa \cite{roberta}; this improves end-to-end performance by roughly +0.6 EM. We believe that parameter sharing among retrieval encoders improves generalizability: because there are more vastly more unique utterances than scenarios, the scenario encoder may overfit to a select set of scenarios, so weight tying enables the joint optimization of both encoders.

\paragraph{Scenario fusion enables better end-to-end modeling.} Because RAF is composed of two neural modules---the retrieval and filling modules---chaining them together arbitrarily may result in information loss. The filling module chiefly uses scenario token embeddings to reason over the retrieval module's outputs. Our results show that scenario fusion, initializing these embeddings using the scenario encoder's final state, improves upon random init by +0.77 EM and +0.65 EM-S.

% \paragraph{Sampling multiple scenario representations avoids overfitting.} Because our retrieval task involves recalling orders of magnitude less items than open-domain tasks (e.g., question answering), our system can easily overfit to particular scenario encodings. Therefore, we explore representation sampling as a means of diversifying the types of encodings our system sees during training. The ablations show sampling multiple representations improves EM by +0.33 compared to using a single representation.

%\begin{itemize}
%    \item Retrieval enables better quality than classification based alternatives
%    \item Hard negatives help and model based negatives are better than edit distance based negatives
%    \item Identity masking for the retrieval loss shows improvements for training
%    \item Scenario fusion further improves the baseline model by allowing scenario embeddings to influence decoding
%    \item Tying parmeters among the encoders furhter improves quality
%    \item Representation sampling allows for further information to be encoded and prevent overfitting to a single scenario representation.
%    \item Retrieval is harder than filling as providing the gold scenario to the model yields near perfect accuracy.
%\end{itemize}

\begin{table*}[t]
\small
\centering
\setlength{\tabcolsep}{4pt}
\begin{tabular}{lrrrrrrrr}
\toprule
 & \multicolumn{8}{c}{Weather Domain (SPIS)} \\
\cmidrule(lr){2-9}
& 0 & 10 & 25 & 50 & 100 & 500 & 1000 & Avg \\
\midrule
RAF$_\textrm{BASE}$ & & & & & & & & \\
\quad Standard Retrieval + Standard Filling & 26.19 & 75.10 & 78.74 & 77.53 & 79.67 & 87.91 & 88.17 & 73.33 \\
\quad Oracle Retrieval + Standard Filling & \textbf{81.68}  & \textbf{90.43} & \textbf{92.13} & \textbf{91.97} & \textbf{93.35} & \textbf{95.74} & \textbf{96.27} & \textbf{91.65} \\
\quad Standard Retrieval + Oracle Filling & 27.67 & 77.84 & 81.70 & 79.92 & 81.78 & 89.88 & 90.44 & 75.60 \\
\bottomrule
\end{tabular}
\caption{Evaluating whether retrieval or filling is the most challenging components of RAF in low-resource settings. We fine-tune several variants of RAF, using either a standard / oracle retriever and a standard / oracle filler, on various SPIS splits of the TOPv2 \cite{chen-2020-topv2} weather domain. RAF with oracle retrieval achieves the best performance, suggesting utterance$\rightarrow$scenario retrieval is the most difficult piece to model.}
\label{tab:gold-low-resource-results2}
\end{table*}

\begin{table*}[t]
\small
\centering
\begin{tabular}{lrrrrrrr}
\toprule
 & \multicolumn{7}{c}{Zero-Shot Evaluation} \\
\cmidrule(lr){3-8}
 & \texttt{en} & \texttt{en$\rightarrow$es} & \texttt{en$\rightarrow$fr} & \texttt{en$\rightarrow$de} & \texttt{en$\rightarrow$hi} & \texttt{en$\rightarrow$th} & Avg \\
\midrule
RAF$_\textrm{BASE}$ & & & & & & & \\
\quad Standard Retrieval + Standard Filling & 81.1 & 56.0 & 54.9 & 40.1 & 32.1 & 30.0 & 42.6 \\
\quad Oracle Retrieval + Standard Filling & \textbf{91.0} & \textbf{68.9} & \textbf{71.6} & \textbf{67.5} & \textbf{47.5} & \textbf{48.9} & \textbf{60.9} \\
\quad Standard Retrieval + Oracle Filling & 83.8 & 66.5 & 62.8 & 45.5 & 40.2 & 46.7 & 52.3 \\
\bottomrule
\end{tabular}
\caption{Evaluating whether retrieval or filling is the most challenging components of RAF in low-resource settings. See Table~\ref{tab:gold-low-resource-results2} for a description of our methodology; we use MTOP \cite{li-2020-mtop} for evaluation instead.}
\label{tab:gold-multilingual-results}
\end{table*}

\subsection{Representation Ablations}

Our results above demonstrate representation sampling is a key piece of end-to-end modeling. But, do we need to include \textit{all} representations outlined in Appendix \ref{appendix:intrinsic_descriptions} in our sampling scheme? The principal reason we use intrinsic representations is for cross-domain robustness, and so we focus our ablation on a low-resource setting, Weather (25 SPIS) in TOPv2-LR. Using the typical low-resource fine-tuning algorithm \cite{chen-2020-topv2}, we adapt multiple versions of RAF to Weather, each omitting a representation (intrinsic automatic, handmade, or examples) from the sampling algorithm.

Table~\ref{tab:intrinsic_ablation} shows these results. \textbf{When comparing the high-level scenario representation, we see that canonical (e.g., ``[IN:GET\_WEATHER [SL:LOCATION ]'') underperforms intrinsic (e.g., ``[ intent | get weather [slot | location ] ]'') by a wide margin (-4.49\%) in the target domain.} This implies RAF \textit{better} understands scenarios' natural language descriptions even if they contain unseen, domain-specific terms. \textbf{Furthermore, we see leveraging all concepts (handmade, automatic, examples) achieves both competitive source and target performance.} Even though excluding handmade improves target performance (+0.53\%), it regresses source performance (-0.28\%), suggesting sampling all representations is more generalizable.

\subsection{Retrieval vs. Filling}

We now turn towards better understanding the aspects our model struggles with. Because RAF jointly optimizes both the retrieval and filling modules, one question we pose is whether the retrieval or filling task is more difficult. We create three versions of RAF: (1) standard retrieval + standard filling, (2) oracle retrieval + standard filling, and (3) standard retrieval + oracle filling. By comparing models (1), (2), and (3), we can judge the relative difficulty of each task.

We begin by evaluating these models in a high-resource setting; on the TOPv2 eval dataset, the standard model gets 87.03\%, retrieval oracle gets 96.56\%, and filling oracle gets 89.34\%. \textbf{Here, the gap between models (1)-(2) is +9.53\%, while the gap between models (1)-(3) is +2.31\%, indicating the retrieval module is the main performance bottleneck.} We also perform experiments in low-resource and multilingual settings, displaying results in Tables \ref{tab:gold-low-resource-results2} and \ref{tab:gold-multilingual-results}, respectively. These results also confirm the same trend: in both settings, the retrieval oracle achieves the best performance, notably achieving +18.3\% averaged across 5 multilingual transfer experiments.

Despite retrieval having the most room for improvement, we also see some evidence filling struggles in certain multilingual transfer cases; for example, providing gold spans can improve en$\rightarrow$th transfer by +16.7\%. As such, there is ample opportunity for optimizing the retrieval and filling modules in future work.

\subsection{Extra-Scenario Generalization}

A core difference between scenario-based and non-scenario-based (seq2seq; autoregressive or non-autoregressive) models is that scenario-based models ``know'' of all scenarios beforehand, while seq2seq models do not, and therefore have to purely rely on generalization. We further quantify the impact that this has by dividing overall EM using two groups: (1) Known vs. Unknown - scenarios in the training dataset vs. test dataset and (2) In-Domain vs. Out-of-Domain - scenarios with a supported intent vs. unsupported intent (e.g., \texttt{IN:UNSUPPORTED\_*}).

From the results in Table~\ref{tab:unknown_frames}, we draw a couple of conclusions. First, RAF outperforms on Unknown EM given that we index unique scenarios across the train, eval, and test datasets before fine-tuning. Second, RAF outperforms on In-Domain EM but underperforms on Out-of-Domain EM. Because RAF leverages intrinsic descriptions of scenarios, the word ``unsupported'' may not precisely capture what it means for an utterance to be in- vs. out-of-domain.

\begin{table*}[h]
\centering
\begin{tabular}{lrrrrr}
\toprule
Model & EM & Known EM & Unknown EM & ID EM & OOD EM\\
\midrule
RAF$_\textrm{BASE}$ & \textbf{87.14} & 88.30 & \textbf{59.96} &  \textbf{88.67} & 44.11 \\
SpanPointer$_\textrm{BASE}$   & 86.76 & \textbf{88.50} & 46.20 &  88.07 & \textbf{49.70} \\
BART$_\textrm{BASE}$  & 86.72 & 88.33 & 49.15 & 88.03 & 49.69 \\
\bottomrule
\end{tabular}
\caption{Comparing EM on Known vs. Unknown and In-Domain (ID) vs. Out-of-Domain (OD) frames. RAF performs better on unknown frames, but struggles with out-of-domain frames.}
\label{tab:unknown_frames}
\end{table*}

\begin{figure*}[h!]
\centering
\includegraphics[width=.9\textwidth]{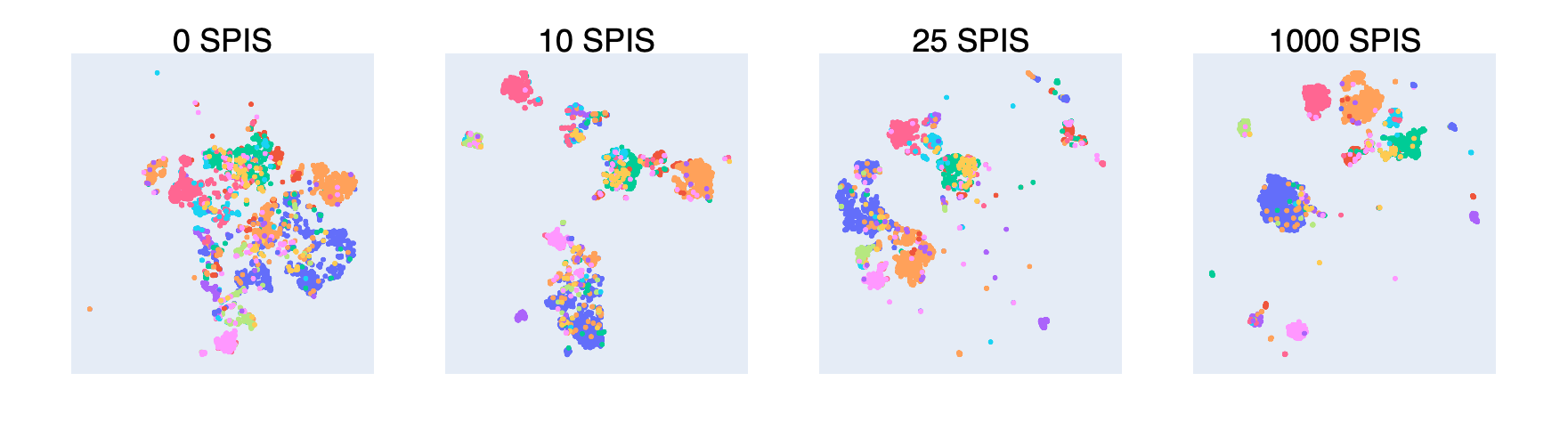}
\caption{Visualizing the semantic space for the weather domain as the model is trained on more and more data. We visualize the index prior to any training (0 SPIS) to low-resource (10, 25 SPIS) and high resource (1000 SPIS). The graphs are TSNE projections of the utterance vectors used for retrieval color coded by the scenario each utterance belongs to.}
\label{fig:scenario}
\end{figure*}

\begin{figure*}[h!]
\centering
\includegraphics[width=\textwidth]{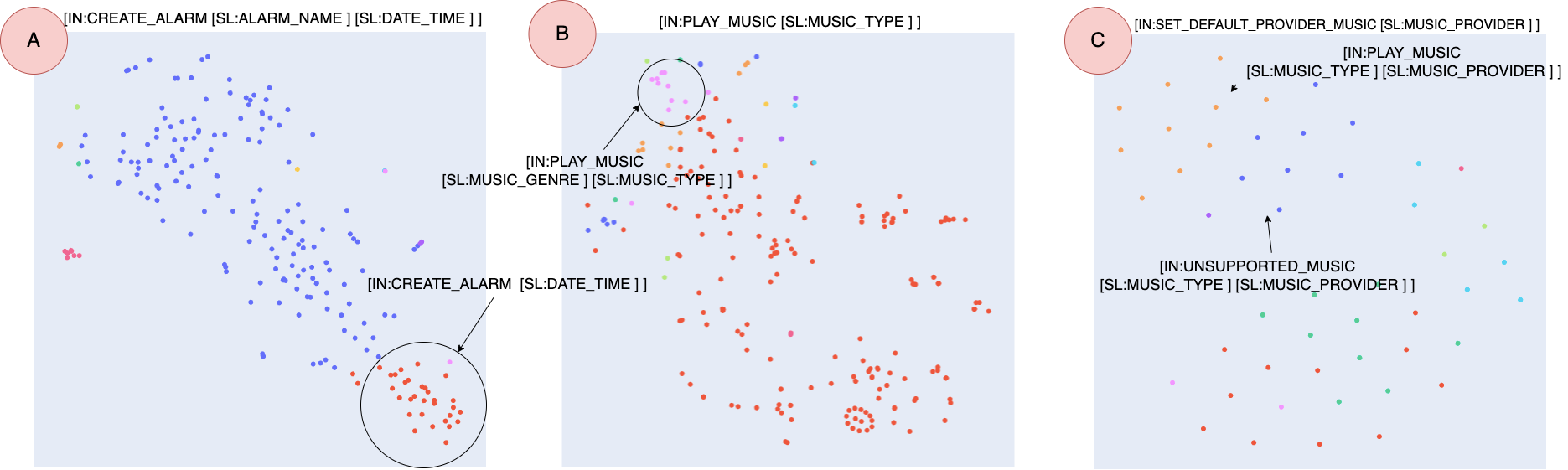}
\caption{Depicting scenario visualizations, where each is a TSNE projection of utterances belonging to the specified scenario color coded by the predicted scenario. (A) covers the scenario \texttt{[IN:CREATE\_ALARM [SL:ALARM\_NAME ] [SL:DATE\_TIME ]} showing a case of ambiguity of alarm names. (B) covers the scenario \texttt{[IN:PLAY\_MUSIC [SL:MUSIC\_TYPE ]} where annotations are missing the slot ``music type''. (C) covers the scenario  \texttt{[IN:SET\_DEFAULT\_PROVIDER\_MUSIC [SL:MUSIC\_PROVIDER ]} showing how our model requires more support here as the predictions for this scenario span OOD and music domain.
} 
\label{fig:debugging}
\end{figure*}

\section{Visualizations}

Because RAF encodes utterances and scenarios into a joint embedding space, we can directly visualize this space to further understand our models' inner-workings. We present two case studies: domain development (\S \ref{sec:domain_dev}) and error analysis (\S \ref{sec:ea}).

\subsection{Domain Development}
\label{sec:domain_dev}

Figure~\ref{fig:scenario} presents an example of performing domain development on the weather domain. Here, we train RAF with 4 dataset sizes (0 SPIS, 10 SPIS, 25 SPIS, and 1,000 SPIS) to simulate zero-shot, few-shot, low-resource, and high-resource settings, respectively. Each utterance (from the high-resource split) is projected using an utterance encoder and colored according to its gold scenario. Interestingly, the zero-shot setting has multiple, apparent clusters, but the overall performance is poor given many scenarios overlap with each other. The clusters spread further apart as the dataset size increases, suggesting the scenarios become more well-defined.

\subsection{Error Analysis}
\label{sec:ea}

While we have demonstrated how RAF refines the utterance-scenario space as we increase dataset size, we now dive deeper into how each space can be used to further analyze domain semantics. In figure~\ref{fig:debugging}, using our high-resource-trained RAF model, we create multiple scenario spaces: each utterance is projected using an utterance encoder and colored according to its predicted frame. We use these scenario spaces in several debugging exercises:

\begin{itemize}
    \item \textbf{Slot Ambiguity:} In Figure \ref{fig:debugging} (a), we investigate the scenario \texttt{[IN:CREATE\_ALARM [SL:ALARM\_NAME ] [SL:DATE\_TIME ]}. Here, we notice a cluster of predictions with the frame \texttt{[IN:CREATE\_ALARM [SL:DATE\_TIME ]} missing the \texttt{[SL:ALARM\_NAME ]}. These map to utterances such as ``I want to wake up at 7 am'' where the annotation has ``wake up'' is \texttt{SL:ALARM\_NAME}; however, our model does not identify this. There are other examples, such as ``wake me up at 7 am'', which are annotated without \texttt{SL:ALARM\_NAME}, leading to ambiguity of whether or not ``wake up'' is an alarm name.
    \item \textbf{Incorrect Annotations:} In Figure \ref{fig:debugging} (b), we investigate the scenario \texttt{[IN:PLAY\_MUSIC [SL:MUSIC\_TYPE ]}. Here, we notice a cluster of predictions with the frame \texttt{[IN:PLAY\_MUSIC [SL:MUSIC\_GENRE ] [SL:MUSIC\_TYPE ]} adding the \texttt{[SL:MUSIC\_GENRE ]}. These map to utterances such as ``Play 1960s music'' where here the annotation only has ``music'' as \texttt{SL:MUSIC\_TYPE}, but our model predicts ``1960s'' as \texttt{SL:MUSIC\_GENRE}. We believe this is an incorrect annotation in this cluster.
    \item \textbf{Underfitting:} In Figure \ref{fig:debugging} (c), we investigate the scenario \texttt{[IN:SET\_DEFAULT\_PROVIDER\_MUSIC [SL:MUSIC\_PROVIDER ]}. This cluster is highly diverse, consisting of predictions from other music intents (e.g., \texttt{IN:PLAY\_MUSIC}) and out-of-domain intents (e.g., \texttt{IN:UNSUPPORTED\_MUSIC}). This scenario may need more data in order to be more properly defined.
\end{itemize}

\section{Related Work}

\subsection{Efficient Decoding for Semantic Parsing}
Recent trends in semantic parsing have started to shift towards the seq2seq paradigm \cite{decoupled, rongali2020don, li-2020-mtop}. However, seq2seq modeling typically has high latency, impeding application in real-world settings. One line of work aims at efficient parsing through logarithmic time decoding; for example, insertion parsing \cite{zhu2020dont}, bottom-up parsing \cite{rubin2021smbop}, and breadth-first parsing \cite{Liu2021X2ParserCA}. Alternative strategies include leveraging insertion transformers to recursively insert ontology tokens into the input, making the model's decoding complexity in the number of intent-slot tokens \cite{RINE}. \citet{nar_semantic_parsing} introduces a one-shot CMLM \cite{ghazvininejad2019maskpredict} -style approach for non-autoregressive semantic parsing, where we model target tokens independently. However, such parses struggle in generalization due to the rigidity of the length prediction tasks. \citet{spanpointernetwork} further augment this with span-based decoding, leading to more consistent length prediction and, as a result, generalizable modeling.

Our work continues in the direction of one-shot decoding by leveraging span-based, non-autoregressive decoding, however, rather than relying on length prediction, we rely on scenario retrieval. We find scenario retrieval as a more interpretable and scalable intermediate task.

\subsection{Scaling Semantic Parsing}

Another critical property of semantic parses lies in reducing data requirements to stand up new domains and scenarios. Existing works rely on leveraging large language models such as BART \cite{lewis2019bart} with augmentations for scaling. In particular \citet{chen-2020-topv2} introduce a meta-learning approach to improve domain scaling in the low-resource setting. Other works such as \cite{Liu2021X2ParserCA, zhu2020dont, RINE} aim to improve scaling through new decoding formulations. \citet{desai2021lowresource} introduce the concept of \textbf{intrinsic modeling} where we provide a human-readable version of the semantic parsing ontology as context to encoding to improve few-shot generalization.

Our work leverages the intrinsic modeling paradigm by building a function $R$ to convert each intent-slot scenario into a readable representation via intent slot descriptions and example priming. Furthermore, our bi-encoder based retrieval setup allows us to inject additional context into each scenario and cache it to an index in order to retain inference efficiency.

\subsection{Retrieval Based Semantic Parsing}
Finally, there has been a recent trend towards dense retrieval in various NLP domains such as machine translation \cite{cai2021neural}, question answering \cite{dpr}, text generation \cite{cai-etal-2019-retrieval} and language modeling \cite{borgeaud2021improving}. 
Recent works also introduce retrieval-based semantic parsing \cite{RetroNLU, pasupat2021controllable}. RetroNLU \cite{RetroNLU} and CASPER \cite{pasupat2021controllable} both leverage a retrieval step to provide examples as context to seq2seq models.

Our approach differs in two ways: (1) We phrase our problem as utterance-to-scenario retrieval rather than utterance-to-utterance retrieval. This allows us to look into supporting new scenarios with minimal-to-no-data required for retrieval. (2) Prior work leverage a separate module \cite{pasupat2021controllable} or separate iteration \cite{RetroNLU} for retrieval. We conduct our retrieval after encoding but prior to decoding as an intermediate step for non-autoregressive parsing. This allows our model to retain similar inference speed to one shot non-autoregressive decoding despite leveraging retrieval.

\section{Conclusion}

In this paper, we tackle scenario-based semantic parsing with retrieve-and-fill (RAF), a coarse-to-fine model which (a) retrieves a scenario with the best alignment to an utterance and (b) fills the scenario with utterance spans in leaf positions. Experiments show our model achieves strong results in high-resource, low-resource, and multilingual settings. The modular nature of our architecture also lends itself well to interpretability and debuggability; we perform several case studies uncovering the inner-workings of our approach.  

%\begin{itemize}
%    \item future work: improve context encoding in scenarios
%    \item An inherent restriction of the RAF and Scenario based parsing approach is knowledge of all possible scenarios. In cases such as compositional parsing, this space can easily explode in an expontential fashion. To mitigate such concerns we can begin exploring semi-autoregressive decoding strategies including multiple passes \cite{ghazvininejad2019maskpredict}, bottom up parsing \cite{rubin2021smbop}, and insertion models \cite{zhu2020dont}.
%\end{itemize}

%\section*{Acknowledgements}

\bibliography{anthology,custom}
\bibliographystyle{acl_natbib}

\newpage
\appendix

\section{Heuristic Negatives}
\label{appendix:heuristic_negative}

In order to understand the importance of model based hard negatives, we develop a simple heuristic to curate a set of hard negative scenarios for each gold scenario. Our heuristic involves selecting the top-N scenarios that share the top level intent but have the lowest Levenshtein edit distance \cite{editdistance} compared to the gold scenario. The full algorithm is described in Algorithm \ref{alg:heuristic_negative}.

In \S \ref{sec:model_ablation} and Table \ref{tab:model_ablation} we present the full results depicting the importance of model based negative sampling. We show that while our heuristic improves on top of no hard negatives (in-batch negatives only), it still lags behind model based hard negatives (-1.37\%).

\begin{algorithm}

\label{euclid}
\begin{algorithmic}[1]
\Procedure{Edit Distance Negatives}{}
\State $\textnormal{S} \gets \text{all scenarios}$
\State $\textnormal{s*} \gets \text{current scenario}$
\State $\textnormal{S}_\textrm{same intent} \gets \text{Scenarios with the same top level intent as Si}$ 
\State $\textnormal{heap} \gets \text{min heap of score and structure}$
\For{$\textit{S}_i \in  \textnormal{S}_\textrm{same intent}$}
\State $\textnormal{score} \gets \textnormal{LevenshteinDistance}(\textnormal{s*},\textit{S}_i) $
\State $\textnormal{heappush}(\textnormal{heap}, \textnormal{score},\textit{S}_i)$
\EndFor
\Return $\text{heap}$
\EndProcedure
\end{algorithmic}
\caption{Heuristic-based negative sampling via edit distance.}
\label{alg:heuristic_negative}
\end{algorithm}

\section{Intrinsic Descriptions}
\label{appendix:intrinsic_descriptions}

In table \ref{tab:repr-types} we provide brief examples of the various representation functions used in RAF training.

\begin{table*}[b]
\centering
\small
\begin{tabular}{ll}
\toprule
$R$ Type & $R$ Example \\
\midrule
\texttt{type-only} & \texttt{[ intent [ slot ] ]} \\
\texttt{automatic-span} & \texttt{[ add time timer [ measurement unit ] ]} \\
\texttt{automatic-type-span} & \texttt{[ intent | add time timer [ slot | measurement unit ] ]} \\
\texttt{automatic-type-span-exs} & \texttt{[ intent | add time timer [ slot | measurement unit | sec / min / hr ] ]} \\
\texttt{curated-span} & \texttt{[ add time to timer [ unit of measurement ] ]} \\
\texttt{curated-type-span} & \texttt{[ intent | add time to timer [ slot | unit of measurement ] ]} \\
\texttt{curated-type-span-exs} & \texttt{[ intent | add time to timer [ slot | unit of measurement | sec / min / hr ] ]} \\
\bottomrule
\end{tabular}
\caption{List of intrinsic representations used for encoding scenarios in RAF.}
\label{tab:repr-types}
\end{table*}

In tables \ref{tab:topv2_intents}, \ref{tab:topv2_slots}, \ref{tab:mtop_intents}, and \ref{tab:mtop_slots} we present the intrinsic hand made descriptions used for each intent/slot on TOPv2 \cite{chen-2020-topv2} and MTOP \cite{li-2020-mtop} respectively. \S \ref{sec:frame_representation} describes the various scenario representations used in full detail.

% In tables \ref{tab:topv2_intents} and \ref{tab:topv2_slots} we present the intrinsic hand made descriptions used for each intent/slot on TOPv2 \cite{chen-2020-topv2} respectively. \S \ref{sec:frame_representation} describes the various scenario representations used in full detail.
\begin{table}[]
\tiny
\begin{tabular}{ll}
\toprule
Intent Token                     & Description                      \\
\midrule
IN:ADD\_TIME\_TIMER              & add time to timer                \\
IN:ADD\_TO\_PLAYLIST\_MUSIC      & add music to playlist            \\
IN:CANCEL\_MESSAGE               & cancel message                   \\
IN:CREATE\_ALARM                 & create alarm                     \\
IN:CREATE\_PLAYLIST\_MUSIC       & create playlist music            \\
IN:CREATE\_REMINDER              & create reminder                  \\
IN:CREATE\_TIMER                 & create timer                     \\
IN:DELETE\_ALARM                 & delete alarm                     \\
IN:DELETE\_REMINDER              & delete reminder                  \\
IN:DELETE\_TIMER                 & delete timer                     \\
IN:DISLIKE\_MUSIC                & dislike music                    \\
IN:GET\_ALARM                    & get alarm                        \\
IN:GET\_BIRTHDAY                 & get birthday                     \\
IN:GET\_CONTACT                  & get contact                      \\
IN:GET\_DIRECTIONS               & get directions                   \\
IN:GET\_DISTANCE                 & get distance between locations   \\
IN:GET\_ESTIMATED\_ARRIVAL       & get estimated arrival time       \\
IN:GET\_ESTIMATED\_DEPARTURE     & get estimated departure time     \\
IN:GET\_ESTIMATED\_DURATION      & get estimated duration of travel \\
IN:GET\_EVENT                    & get event                        \\
IN:GET\_EVENT\_ATTENDEE          & get event attendee               \\
IN:GET\_EVENT\_ATTENDEE\_AMOUNT  & get amount of event attendees    \\
IN:GET\_EVENT\_ORGANIZER         & get organizer of event           \\
IN:GET\_INFO\_CONTACT            & get info of contact              \\
IN:GET\_INFO\_ROAD\_CONDITION    & get info of road condition       \\
IN:GET\_INFO\_ROUTE              & get info of route                \\
IN:GET\_INFO\_TRAFFIC            & get info of traffic              \\
IN:GET\_LOCATION                 & get location                     \\
IN:GET\_LOCATION\_HOME           & get location of my home          \\
IN:GET\_LOCATION\_HOMETOWN       & get location of my hometown      \\
IN:GET\_LOCATION\_SCHOOL         & get location of school           \\
IN:GET\_LOCATION\_WORK           & get location of work             \\
IN:GET\_MESSAGE                  & get message                      \\
IN:GET\_RECURRING\_DATE\_TIME    & get recurring date or time       \\
IN:GET\_REMINDER                 & get reminder                     \\
IN:GET\_REMINDER\_AMOUNT         & get amount of reminders          \\
IN:GET\_REMINDER\_DATE\_TIME     & get date or time of reminder     \\
IN:GET\_REMINDER\_LOCATION       & get location of reminder         \\
IN:GET\_SUNRISE                  & get info of sunrise              \\
IN:GET\_SUNSET                   & get info of sunset               \\
IN:GET\_TIME                     & get time                         \\
IN:GET\_TIMER                    & get timer                        \\
IN:GET\_TODO                     & get todo item                    \\
IN:GET\_WEATHER                  & get weather                      \\
IN:HELP\_REMINDER                & get help reminder                \\
IN:IGNORE\_MESSAGE               & ignore message                   \\
IN:LIKE\_MUSIC                   & like music                       \\
IN:LOOP\_MUSIC                   & loop music                       \\
IN:NEGATION                      & negate                           \\
IN:PAUSE\_MUSIC                  & pause music                      \\
IN:PAUSE\_TIMER                  & pause timer                      \\
IN:PLAY\_MUSIC                   & play music                       \\
IN:PREVIOUS\_TRACK\_MUSIC        & play previous music track        \\
IN:REACT\_MESSAGE                & react to message                 \\
IN:REMOVE\_FROM\_PLAYLIST\_MUSIC & remove from music playlist       \\
IN:REPLAY\_MUSIC                 & replay music                     \\
IN:PREVIOUS\_TRACK\_MUSIC        & play previous music track        \\
IN:REACT\_MESSAGE                & react to message                 \\
IN:REMOVE\_FROM\_PLAYLIST\_MUSIC & remove from music playlist       \\
IN:REPLAY\_MUSIC                 & replay music                     \\
IN:REPLY\_MESSAGE                & reply to message                 \\
IN:RESTART\_TIMER                & restart timer                    \\
IN:RESUME\_TIMER                 & resume timer                     \\
IN:SELECT\_ITEM                  & select item                      \\
IN:SEND\_MESSAGE                 & send message                     \\
IN:SEND\_TEXT\_MESSAGE           & send text message                \\
IN:SET\_DEFAULT\_PROVIDER\_MUSIC & set default music provider       \\
IN:SILENCE\_ALARM                & silence alarm                    \\
IN:SKIP\_TRACK\_MUSIC            & skip music track                 \\
IN:SNOOZE\_ALARM                 & snooze alarm                     \\
IN:START\_SHUFFLE\_MUSIC         & start shuffling music            \\
IN:STOP\_MUSIC                   & stop music                       \\
IN:SUBTRACT\_TIME\_TIMER         & subtract time from timer         \\
IN:UNSUPPORTED\_ALARM            & unsupported alarm request        \\
IN:UNSUPPORTED\_EVENT            & unsupported event request        \\
IN:UNSUPPORTED\_MESSAGING        & unsupported messaging request    \\
IN:UNSUPPORTED\_MUSIC            & unsupported music request        \\
IN:UNSUPPORTED\_NAVIGATION       & unsupported navigation request   \\
IN:UNSUPPORTED\_TIMER            & unsupported timer request        \\
IN:UNSUPPORTED\_WEATHER          & unsupported weather request      \\
IN:UPDATE\_ALARM                 & update alarm                     \\
IN:UPDATE\_DIRECTIONS            & update directions                \\
IN:UPDATE\_REMINDER              & update reminder                  \\
IN:UPDATE\_REMINDER\_DATE\_TIMER & update date time of reminder     \\
IN:UPDATE\_REMINDER\_TODO        & update todo of reminder          \\
IN:UPDATE\_TIMER                 & update timer \\
\bottomrule
\end{tabular}
\caption{List of intrinsic handmade descriptions for intents in TOPv2 \cite{chen-2020-topv2}.}
\label{tab:topv2_intents}
\end{table}

\begin{table}[]
\tiny
\begin{tabular}{ll}
\toprule
Slot Token                     & Description                   \\
\midrule
SL:AGE                         & age of person                 \\
SL:ALARM\_NAME                 & alarm name                    \\
SL:AMOUNT                      & amount                        \\
SL:ATTENDEE                    & attendee                      \\
SL:ATTENDEE\_ADDED             & attendee to be added          \\
SL:ATTENDEE\_EVENT             & attendee of event             \\
SL:ATTENDEE\_REMOVED           & attendee to be removed        \\
SL:ATTRIBUTE\_EVENT            & attribute of event            \\
SL:BIRTHDAY                    & birthday                      \\
SL:CATEGORY\_EVENT             & category of event             \\
SL:CATEGORY\_LOCATION          & category of location          \\
SL:CONTACT                     & contact                        \\
SL:CONTACT\_RELATED            & contact related               \\
SL:CONTENT\_EMOJI              & content text with emoji       \\
SL:CONTEnT\_EXACT              & content text                  \\
SL:DATE\_TIME                  & date or time                  \\
SL:DATE\_TIME\_ARRIVAL         & date or time of arrival       \\
SL:DATE\_TIME\_BIRTHDAY        & date or time of birthday      \\
SL:DATE\_TIME\_DEPARTURE       & date or time of departure     \\
SL:DATE\_TIME\_NEW             & new date or time              \\
SL:DATE\_TIME\_RECURRING       & recurring date or time        \\
SL:DESTINATION                 & travel destination            \\
SL:DURATION                    & duration                      \\
SL:FREQUENCY                   & frequency                     \\
SL:GROUP                       & group                         \\
SL:JOB                         & job                           \\
SL:LOCATION                    & location                      \\
SL:LOCATION\_CURRENT           & current location              \\
SL:LOCATION\_HOME              & location of my home           \\
SL:LOCATION\_MODIFIER          & location modifier             \\
SL:LOCATION\_USER              & location of user              \\
SL:LOCATION\_WORK              & location of work              \\
SL:MEASUREMENT\_UNIT           & unit of measurement           \\
SL:METHOD\_RETRIEVAL\_REMINDER & method of retrieving reminder \\
SL:METHOD\_TIMER               & method of timer               \\
SL:METHOD\_TRAVEL              & method of traveling           \\
SL:MUSIC\_ALBUM\_TITLE         & title of music album          \\
SL:MUSIC\_ARIST\_NAME          & name of music artist          \\
SL:MUSIC\_GENRE                & genre of music                \\
SL:MUSIC\_PLAYLIST\_TITLE      & title of music playlist       \\
SL:MUSIC\_PROVIDER\_NAME       & name of music provider        \\
SL:MUSIC\_RADIO\_ID            & id of music radio             \\
SL:MUSIC\_TRACK\_TITLE         & title of music track          \\
SL:MUSIC\_TYPE                 & type of music                 \\
SL:MUTUAL\_EMPLOYER            & mutual employer               \\
SL:MUTUAL\_LOCATION            & mutual location               \\
SL:MUTUAL\_SCHOOL              & mutual school                 \\
SL:NAME\_APP                   & name of app                   \\
SL:NAME\_EVENT                 & name of event                 \\
SL:OBSTRUCTION\_AVOID          & obstruction to avoid          \\
SL:ORDINAL                     & ordinal                       \\
SL:ORGANIZER\_EVENT            & obstruction to avoid          \\
SL:ORDINAL                     & ordinal                       \\
SL:ORGANIZER\_EVENT            & organizer of event            \\
SL:PATH                        & path                          \\
SL:PATH\_AVOID                 & path to avoid                 \\
SL:PERIOD                      & time period                   \\
SL:PERSON\_REMINDED            & person to be reminded         \\
SL:PERSON\_REMINDED\_ADDED     & added person to be reminded   \\
SL:PERSON\_REMINDED\_REMOVED   & removed person to be reminded \\
SL:POINT\_ON\_MAP              & point on map                  \\
SL:RECIPIENT                   & message recipient             \\
SL:RECURRING\_DATE\_TIME       & recurring date or time        \\
SL:RECURRING\_DATE\_TIME\_NEW  & new recurring date or time    \\
SL:RESOURCE                    & resource                      \\
SL:ROAD\_CONDITION             & road condition                \\
SL:ROAD\_CONDITION\_AVOID      & road condition to avoid       \\
SL:SEARCH\_RADIUS              & search radius                 \\
SL:SENDER                      & message sender                \\
SL:SOURCE                      & travel source                 \\
SL:TAG\_MESSGE                 & tag of message                \\
SL:TIMER\_NAME                 & timer name                    \\
SL:TIME\_ZONE                  & time zone                     \\
SL:TODO                        & todo item                     \\
SL:TODO\_NEW                   & new todo item                 \\
SL:TYPE\_CONTACT               & contact type                  \\
SL:TYPE\_CONTENT               & content type                  \\
SL:TYPE\_INFO                  & info type                     \\
SL:TYPE\_REACTION              & reaction type                 \\
SL:TYPE\_RELATION              & relation type                 \\
SL:UNIT\_DISTANCE              & unit of distance              \\
SL:WAYPOINT                    & waypoint                      \\
SL:WAYPOINT\_ADDED             & waypoint to be added          \\
SL:WAYPOINT\_AVOID             & waypoint to avoid             \\
SL:WEATHER\_ATTRIBUTE          & weather attribute             \\
SL:WEATHER\_TEMPERATURE\_UNIT  & unit of temperature \\
\bottomrule
\end{tabular}
\caption{List of intrinsic handmade descriptions for slots in TOPv2 \cite{chen-2020-topv2}.}
\label{tab:topv2_slots}
\end{table}

\begin{table}[]
\tiny
\begin{tabular}{ll}
\toprule
Intent Token                     & Description                      \\
\midrule
IN:FOLLOW\_MUSIC              & follow music  \\
IN:GET\_JOB              & get job  \\
IN:GET\_GENDER              & get gender  \\
IN:GET\_UNDERGRAD              & get undergrad education  \\
IN:GET\_MAJOR              & get college major  \\
IN:DELETE\_PLAYLIST\_MUSIC              & delete playlist  \\
IN:GET\_EDUCATION\_DEGREE              & get education degree of person  \\
IN:GET\_AGE              & get age of person  \\
IN:DISPREFER              & dislike item  \\
IN:RESUME\_MUSIC              & resume music  \\
IN:QUESTION\_MUSIC              & question about music  \\
IN:CREATE\_CALL              & create a caoo  \\
IN:GET\_AIRQUALITY              & get airquality  \\
IN:GET\_CALL\_CONTACT              & get contact for caller  \\
IN:SET\_UNAVAILABLE              & set status to unavailable  \\
IN:END\_CALL              & end call  \\
IN:STOP\_SHUFFLE\_MUSIC              & stop shuffle of music  \\
IN:PREFER              & prefer item  \\
IN:GET\_LANGUAGE              & get language  \\
IN:SET\_AVAILABLE              & set available  \\
IN:GET\_GROUP              & get group  \\
IN:ANSWER\_CALL              & answer call  \\
IN:GET\_CONTACT\_METHOD              & get method to contact  \\
IN:UPDATE\_METHOD\_CALL              & update method of call  \\
IN:GET\_ATTENDEE\_EVENT              & get attendee for event  \\
IN:UPDATE\_CALL              & update call  \\
IN:GET\_LIFE\_EVENT              & get life event  \\
IN:REPEAT\_ALL\_MUSIC              & repeat all music  \\
IN:GET\_EDUCATION\_TIME              & get education time  \\
IN:QUESTION\_NEWS              & question about news  \\
IN:GET\_EMPLOYER              & get employer  \\
IN:IGNORE\_CALL              & ignore call  \\
IN:REPEAT\_ALL\_OFF\_MUSIC              & turn of repeat  \\
IN:UNLOOP\_MUSIC              & turn loop off  \\
IN:SET\_DEFAULT\_PROVIDER\_CALLING              & set default provider for calling  \\
IN:GET\_AVAILABILITY              & get avalability of contact  \\
IN:HOLD\_CALL              & hold call  \\
IN:GET\_LIFE\_EVENT\_TIME              & get time of life event  \\
IN:SHARE\_EVENT              & share event  \\
IN:CANCEL\_CALL              & cancel call  \\
IN:SET\_RSVP\_YES              & set rsvp to yes  \\
IN:PLAY\_MEDIA              & play media  \\
IN:GET\_TRACK\_INFO\_MUSIC              & get information about the current track  \\
IN:GET\_DATE\_TIME\_EVENT              & get the date time of the event  \\
IN:SET\_RSVP\_NO              & set rsvp to no  \\
IN:MERGE\_CALL              & marge call  \\
IN:UPDATE\_REMINDER\_LOCATION              & update the location of the reminder  \\
IN:GET\_MUTUAL\_FRIENDS              & get mutual friends  \\
IN:GET\_MESSAGE\_CONTACT              & get information about message contact  \\
IN:GET\_LYRICS\_MUSIC              & get lyrics about the song  \\
IN:GET\_INFO\_RECIPES              & get information about recipe  \\
IN:GET\_DETAILS\_NEWS              & get news details  \\
IN:GET\_EMPLOYMENT\_TIME              & get employment time  \\
IN:GET\_RECIPES              & get a recipe  \\
IN:GET\_CALL              & get call  \\
IN:GET\_CALL\_TIME              & get time of the call  \\
IN:GET\_CATEGORY\_EVENT              & get the category of the event  \\
IN:RESUME\_CALL              & resume the call  \\
IN:IS\_TRUE\_RECIPES              & ask question about recipes  \\
IN:SET\_RSVP\_INTERESTED              & set rsvp to interested  \\
IN:GET\_STORIES\_NEWS              & get news stories  \\
IN:SWITCH\_CALL              & switch call  \\
IN:REWIND\_MUSIC              & rewind the song  \\
IN:FAST\_FORWARD\_MUSIC              & forward the song  \\
\bottomrule
\end{tabular}
\caption{List of intrinsic handmade descriptions for intents in MTOP \cite{li-2020-mtop}.}
\label{tab:mtop_intents}
\end{table}

\begin{table}[]
\tiny
\begin{tabular}{ll}
\toprule
Slot Token                     & Description                      \\
\midrule
SL:GENDER              & gender of person  \\
SL:RECIPES\_TIME\_PREPARATION              & time to prepare recipe  \\
SL:RECIPES\_EXCLUDED\_INGREDIENT              & exclude ingredient for recipe  \\
SL:USER\_ATTENDEE\_EVENT              & attendee of event  \\
SL:MAJOR              & major  \\
SL:RECIPES\_TYPE              & type of recipe  \\
SL:SCHOOL              & school  \\
SL:TITLE\_EVENT              & title of event  \\
SL:MUSIC\_ALBUM\_MODIFIER              & type of album  \\
SL:RECIPES\_DISH              & recipe dish  \\
SL:NEWS\_TYPE              & type of news  \\
SL:RECIPES\_SOURCE              & source of recipe  \\
SL:RECIPES\_DIET              & diet of recipe  \\
SL:RECIPES\_UNIT\_NUTRITION              & nutrition unit of recipe  \\
SL:MUSIC\_REWIND\_TIME              & time to rewind music  \\
SL:RECIPES\_TYPE\_NUTRITION              & nutrition type of recipe  \\
SL:CONTACT\_METHOD              & method to contact  \\
SL:SIMILARITY              & similarity  \\
SL:PHONE\_NUMBER              & phone number  \\
SL:NEWS\_CATEGORY              & category of news  \\
SL:RECIPES\_INCLUDED\_INGREDIENT              & ingredient in recipe  \\
SL:EDUCATION\_DEGREE              & education degree  \\
SL:RECIPES\_RATING              & rating of recipe  \\
SL:CONTACT\_REMOVED              & removed contact  \\
SL:NEWS\_REFERENCE              & news reference  \\
SL:METHOD\_RECIPES              & method of recipe  \\
SL:LIFE\_EVENT              & life event  \\
SL:RECIPES\_MEAL              & recipe meal  \\
SL:NEWS\_TOPIC              & news topic  \\
SL:RECIPES\_ATTRIBUTE              & recipe attribute  \\
SL:EMPLOYER              & employer  \\
SL:RECIPES\_COOKING\_METHOD              & cooking method of recipe  \\
SL:RECIPES\_CUISINE              & cuisine of recipe  \\
SL:MUSIC\_PLAYLIST\_MODIFIER              & music playlist modifier  \\
SL:RECIPES\_QUALIFIER\_NUTRITION              & nutrition qualifier of recipe  \\
SL:METHOD\_MESSAGE              & method to send message  \\
SL:RECIPES\_UNIT\_MEASUREMENT              & unit of measurement in recipe  \\
SL:CONTACT\_ADDED              & added contact  \\
SL:NEWS\_SOURCE              & news source  \\
\bottomrule
\end{tabular}
\caption{List of intrinsic handmade descriptions for slots in MTOP \cite{li-2020-mtop}.}
\label{tab:mtop_slots}
\end{table}

\section{Hyperparameters}
\label{appendix:hyperparameters}
In this section we describe the hyper parameters for training our various RAF models.

\paragraph{Architecture Parameters.}
For our RAF architectures we leverage a shared RoBERTa \cite{roberta} or XLM-R \cite{conneau-2020-xlmr} encoder for both the utterance and scenario encoders. We augment each of these encoders with an additional projection layer with a hidden dimension of 768 (base models) or 1024 (large models). For our span pointer decoder we leverage a 1 layer transformer decoder with the same hidden dimension as the respective encoder (1L, 768/1024H, 16/24A).

\paragraph{Optimization Parameters.}
We train our models with the Adam \cite{kingma2014adam} optimizer along with a warmup and linear decay. We train our models across 8 GPUs with 32GB memory each. Additionally we optionally augment our models with the R3F loss \cite{RXF} based on validation set tuning in each setting. To determine hyperparameters, we conduct hyperparameter sweeps with 56 iterations each. The hyperparameters for the high resource runs on TOPv2 \cite{chen-2020-topv2} are described in Table \ref{table:hp_raf}.

\begin{table*}[h]
\small
\centering
\begin{tabular}{@{}lrrrrrr@{}}
\toprule
\multirow{2}{*}{Parameter} & \multicolumn{3}{c}{RAF Models}& \\
\cmidrule(lr){2-5}
 & RoBERTa$_\textrm{BASE}$ & RoBERTa$_\textrm{LARGE}$ & XLM-R$_\textrm{BASE}$  \\ \midrule
Epochs &  \multicolumn{3}{c}{40}  \\
Optimizer &  \multicolumn{3}{c}{Adam}  \\
Weight Decay & \multicolumn{3}{c}{0.01} \\
$\epsilon$ & \multicolumn{3}{c}{1e-8} \\
Warmup Period (steps) &  \multicolumn{3}{c}{1000} \\
Learning Rate Scheduler &  \multicolumn{3}{c}{Linear Decay} \\
Learning Rate & 0.00002 & 0.00003 & 0.00002 \\
Batch Size & 40 & 12 & 16  \\
$\beta \mathcal{L}_\textrm{Retrieval}$ & 2.69 & 4 & 2.69  \\
$\mathcal{L}_\textrm{filling} \textrm{Label Smoothing Penalty}$ & \multicolumn{3}{c}{0.2} \\
\# GPU & \multicolumn{3}{c}{8} \\
GPU Memory & 32GB  & 32GB  & 32GB \\
\bottomrule
\end{tabular}
\caption{Hyperparameter values for RAF architectures.}
\label{table:hp_raf}
\end{table*}

\end{document}